\documentclass[11pt,letterpaper]{article}
\usepackage{acl2017}
\usepackage{times}
\usepackage{latexsym}

\usepackage{amsmath}
\usepackage{amsfonts}
\usepackage{graphicx}
\usepackage{algorithm}
\usepackage{algorithmic}
\usepackage{multirow}
\usepackage{rotating}
\usepackage{xcolor}
\usepackage[footnotesize]{caption}
\usepackage{framed}
\usepackage{scrextend}
\usepackage{booktabs}
\usepackage{paralist}
\usepackage[caption=false]{subfig}
\aclfinalcopy

\def\emnlppaperid{***}

\title{Simple Open Stance Classification for Rumour Analysis}

\author{Ahmet Aker\textsuperscript{a,b} \and Leon Derczynski\textsuperscript{a} \and Kalina Bontcheva\textsuperscript{a} \\
        Department of Computer Science, University of Sheffield\textsuperscript{a}\\
        Department of Information Engineering, University of Duisburg-Essen\textsuperscript{b}\\
        {\tt a.aker@is.inf.uni-due.de, leon.derczynski@sheffield.ac.uk}\\
        {\tt K.Bontcheva@sheffield.ac.uk}}
\date{}
\hypersetup{draft} 
\begin{document}

\maketitle

\begin{abstract}
Stance classification determines the attitude, or stance, in a (typically short) text. The task has powerful applications, such as the detection of fake news or the automatic extraction of attitudes toward entities or events in the media. This paper describes a surprisingly simple and efficient classification approach to open stance classification in Twitter, for rumour and veracity classification. The approach profits from a novel set of automatically identifiable problem-specific features, which significantly boost classifier accuracy and achieve above state-of-the-art results on recent benchmark datasets. This calls into question the value of using complex sophisticated models for stance classification without first doing informed feature extraction.
%Our results show that the features we intriduce add additional boost and lead to state-of-the-art figures for the recent benchmark datasets. 
%features to do stance classification, evaluated over recent benchmark datasets.
%Unlike many prior approaches, we use a surprisingly simple classification approaches with features reported from related work but also introduce novel features to further boost the accuracy figures and to achieve state-of-the-art results.
%Using simple classification approaches and features reported by related work we can already achieve results  that outperform current systems on stance detection. 

\end{abstract}

\section{Introduction}

Stance detection is the problem of classifying the attitude taken by an author in a short piece of text. Typical stances include showing support, denying, commenting on or querying an existing claim or fact. Knowing the stance that authors hold in response to claims, e.g. in online commentary, gives useful insights.
It can reveal rumours and fake news claims as the discourse around them is monitored~\cite{procter2013reading}. Stance reflects how certain authors are of a claim's veracity~\cite{biber2006stance}, which enables the effective detection of potential false rumours~\cite{lukasik2015classifying}. Stance also reveals how online populations react to business and political news.

This paper addresses the general-purpose, or {\em open} stance classification task.
This is distinct from {\em target-specific} stance classification, as in~\newcite{augenstein-EtAl:2016:EMNLP2016} and~\newcite{mohammad2016semeval}, which focus on stances towards known, pre-determined targets.
In the latter task, the target has already been extracted, from e.g. conversational cues.
Target-specific stance classification is suited to situations where the target is already known, such as analyses of a specific product or political actor.
In contrast, the open stance classification task is appropriate in emerging news or novel contexts, such as working with online media or streaming news analysis.

%KLB: explain in the paper the relationship between the PHEME datasets (used by Michal et al.) and RumourEval

%KLB: "in fact, in that same EMNLP paper [Augenstein 2016] I had constructed an argument on the difference between the PHEME task and the target-specific stance for entities"

%Prior general stance classification: arkaitz~\cite{zubiaga2016stance}, michal~\cite{lukasik2015classifying}

Open stance classification is often applied in rumour resolution.
Since attitudes in discourse around a claim are indicative not only of the controversiality of the claim, but also can act as a proxy for its veracity, it is reasonable to consider the application of open stance detection for rumour analysis.
Indeed, many approaches to rumour and fake news analysis rely on this signal~\cite{rumoureval2017}\footnote{http://approximatelycorrect.com/2017/01/23/is-fake-news-a-machine-learning-problem/}.
In veracity analysis, the claim is already known, and the goal is to gather observations and analyse crowd reaction in order to resolve the claim.
Instead of being concerned with specific targets, we apply non-targeted -- {\em open} -- stance analysis to messages replying to a claim, where the target may vary but the high-level rumour topic rumour remains the same.

Our simple approach to open stance classification implements common features used in stance classification reported by related work (e.g. bag-of-words, named entities, user activity information, URL presence). We extend this with problem-specific features (which we refer to as the AF features) designed to capture how users react to tweets and express confidence in them. Our results show adding these features gives significantly higher performance on benchmark datasets, compared to recent state-of-the-art systems. We make our classifier freely available on the PHEME software repository.\footnote{\url{https://gate.ac.uk/wiki/pheme-stance.html}} 

%Furthermore, omitting the AF features proposed in this work leads to performance loss and struggle to compete with current work applied on the recent SemEval workbench data. 

The outline of the paper is as follows. First we describe related work (Section \ref{sec:relatedWork}) and then introduce our method along with the classification techniques used and features extracted (Section \ref{sec:method}). Next, Section \ref{sec:exSettings} describes our experimental set-ups, followed by results in Section \ref{sec:results}. We report on feature analysis in Section \ref{sec:analysis}, prior to concluding the paper (Section \ref{sec:conclusion}). 

\section{Related Work}
\label{sec:relatedWork}

The first study that tackles automatic stance classification is that of Qazvinian et al. \shortcite{qazvinian2011rumor}. With a dataset containing 10K tweets and using a Bayesian classifier and three types of features categorised as ``content'', ``network'' and ``Twitter specific memes'', the authors achieved an accuracy of 93.5\%. Similar to them, Hamidian and Diab \shortcite{hamidian2015rumor} perform rumour stance classification by applying supervised machine learning using the dataset created by Qazvinian et al. \shortcite{qazvinian2011rumor}. However, instead of Bayesian classifiers, the authors use J48 decision tree implemented within the Weka platform \cite{hall2009weka}. The features from Qazvinian et al. \shortcite{qazvinian2011rumor} are adopted and extended with time-related information and the hastags themselves, instead of the content of the hashtag as used by Qazvinian et al. \shortcite{qazvinian2011rumor}. In addition to the feature categories introduced above, Hamidian and Diab \shortcite{hamidian2015rumor} introduce another feature category, namely ``pragramatic''. The pragmatic features include named entity, event, sentiment and emoticons. The evaluation of the performance is casted as either 1-step problem containing a 6 class classification task (not rumour, 4 classes of stance and not determined by the annotator) or 2-step problem containing first a 3 class classification task (non-rumour, rumour, and not determined), followed by a 4 class classification task (stance classification). The two step approach achieves better performance with 82.9\% F-1 measure, compared to 74\% with the 1-step approach. The authors also report that the best performing features were the content based features and the worst performing ones -- the network and Twitter specific features. In their most recent paper, Hamidian and Diab \shortcite{hamidian2016rumor} introduce the Tweet Latent Vector (TLV) approach that is obtained by applying the Semantic Textual Similarity model proposed by Guo and Diab \shortcite{guo2012modeling}. The authors compare the TLV approach to their own earlier system, as well as to the original features of Qazvinian et al. \shortcite{qazvinian2011rumor} and show that the TLV approach outperforms both baselines.

Liu et al. \shortcite{liu2015real} use a rule-based method and show that it outperforms the approach reported by Qazvinian et al. \shortcite{qazvinian2011rumor}. Zeng et al. \shortcite{zeng2016unconfirmed} enrich the feature sets investigated by earlier studies by features derived from the Linguistic Inquiry and Word Count (LIWC) dictionaries \cite{tausczik2010psychological}. Lukasik et al. \shortcite{lukasik2016hawkes} investigate Gaussian Processes as rumour stance classifier. For the first time the authors also use Brown Clusters to extract the features for each tweet. Unlike researchers above, Lukasik et al. evalute on the rumour data released by Zubiaga et al. \shortcite{zubiaga2016analysing}, where they report an accuracy of 67.7\%. This result is achieved when the classifier is trained on $n-1$ rumours and tested on the n$^{th}$ rumour. However, the authors achieve substantially better results when a small proportion of the in-domain data (data from the n$^{th}$ rumour) is also included in the training (68.6\% accuracy). Performance scores differ substantially from those in the studies described above, given that Lukasik et al. \shortcite{lukasik2016hawkes} tackled classification of stance in new rumours that differ from those in the training set.

Subsequent work has also tackled stance classification for new, unseen rumours. Zubiaga et al. \cite{zubiaga2016stance} moved away from the classification of tweets in isolation, focusing instead on Twitter 'conversations' \cite{tolmie2015microblog} initiated by rumours, as part of the {\sc Pheme} project~\cite{derczynski2014pheme}. They looked at tree-structured conversations initiated by a rumour and followed by tweets responding to it by supporting, denying, querying or commenting on the rumour.

Rumour stance classification for tree structured conversations has also been studied in the RumourEval shared task at SemEval 2017 \cite{rumoureval2017}. Subtask A there consisted of stance classification of individual tweets discussing a rumour within a conversational thread as one of \textit{support}, \textit{deny}, \textit{query}, or \textit{comment}. Eight participanting teams submitted results to this task. Most of the systems viewed this task as a 4-way single tweet classification task, with the exception of the best performing system by Kochkina et al. \shortcite{kochkina2017turing}, as well as the systems by Wang et al. \shortcite{wang2017ecnu} and Singh et al. \shortcite{singh2017iitp}. The winning system addressed the task as a sequential classification problem, where the stance of each tweet takes into consideration the features and labels of the preceding tweets. The system by Singh et al. \shortcite{singh2017iitp} takes as input pairs of source and reply tweets, whereas Wang et al. \shortcite{wang2017ecnu} addressed class imbalance by decomposing the problem into a two step classification task -- first distinguishing between comments and non-comments and then classifying non-comment tweets as one of support, deny or query. Half of the systems employed ensemble classifiers, where classification was obtained through majority voting \cite{wang2017ecnu,garcialozano2017mama,bahuleyan2017uwaterloo,srivastava2017dfki}. In some cases the ensembles were hybrid, consisting both of machine learning classifiers and manually created rules, with differential weighting of classifiers for different class labels \cite{wang2017ecnu,garcialozano2017mama,srivastava2017dfki}. Three systems used deep learning, with Kochkina et al. \shortcite{kochkina2017turing} employing LSTMs for sequential classification, Chen et al. \shortcite{chen2017ikm} using convolutional neural networks (CNN) for obtaining the representation of each tweet, assigned a probability for a class by a softmax classifier and Garc\'ia Lozano et al. \shortcite{garcialozano2017mama} using CNN as one of the classifiers in their hybrid conglomeration. The remaining two systems by Enayet and El-Beltagy \shortcite{enayet2017niletmrg} and Singh et al. \shortcite{singh2017iitp} used support vector machines with a linear and polynomial kernel respectively. 

%Unlike related work we 

%test all reported machine learning techniques on the same data set. This helps to compare their performance better. In addition, we evaluate the best performing model using out-of-domain data. This gives reliable indication about how portable a model is when used in an unseen environment. 

\section{Method}
\label{sec:method}

\subsection{Data}

In our experiments we used two different data sets: RumourEval dataset~\cite{rumoureval2017} and the PHEME dataset \cite{zubiaga2016analysing}. In the PHEME dataset the authors identify rumours associated with events, collect conversations sparked by those rumours in the form of replies and annotate each of the tweets in the conversations for stance. These data consist of tweets from 5 different events: Ottawa shooting, Ferguson riots, Germanwings crash, Charlie Hebdo and Sydney siege. Each dataset has a different number of rumours where each rumour contains tweets marked with stance annotations: ``supporting'', ``questioning'', ``denying'' or ``commenting''. A summary of the data is given in Table \ref{table:DatacountsPheme}.

\begin{table}[t]
\centering
\footnotesize
\begin{tabular}{lrrrrr}
\toprule
Dataset & Rumours & S & D & Q & C \\
\midrule 
Ottawa shooting & 58 & 161 & 76 & 64 & 481\\
Ferguson riots& 46 & 192 & 83 & 94 & 685\\
%Germanwings crash & 68 & 177 & 12 & 28 & 169\\
Charlie Hebdo & 74 & 236 & 56 & 51 & 710\\
Sydney siege & 71 & 89 & 4 & 99 & 713\\
\bottomrule 
\end{tabular}
\caption{PHEME Data: Counts of tweets with supporting (S), denying (D), questioning (Q) and commenting (C) labels in each event collection.}
\label{table:DatacountsPheme}
\end{table}

The RumourEval dataset is derived from the PHEME dataset, however, for the purpose of the RumourEval shared Task A the data has a given split into training and testing.
This provides an established basis for evaluation.
The training data draws from stories in 2014--2016, from the earlier PHEME dataset.
The evaluation split covers two new stories, both from 2016: first, the disappearance of Marina Joyce, a British Youtube personality, who was rumoured to have been abducted in July 2016.
There was significant speculation in social media, and the case was brought to a concrete resolution as the police investigated and posted an open public response.
The second story was that Hillary Clinton had pneumonia during mid-September 2016.
The prevalence and spread of this story could be tracked easily, and it emerged in a short space of time, though among background noise of speculative, unsubstantiated claims about her and her opponent's health.
More details about this dataset can be obtained from the SemEval website\footnote{http://alt.qcri.org/semeval2017/task8/}. 

In keeping with prior work \cite{zeng2016unconfirmed,lukasik2016hawkes,zubiaga2016stance}, our experiments assume that incoming tweets already belong to a particular rumour, e.g. a user is tracking tweets related to a certain rumour. For each new tweet, features are extracted into a feature vector, which is then used to assign each tweet its stance towards the rumour.

\subsection{Classifiers}
\label{sec:classifiers}

We experiment with three different, well known machine learning classifiers: (1) a decision tree, J48; (2) Random Forests~\cite{breiman2001random}; and (3) an Instance Based classifier (K-NN). 
%All are run using Weka \cite{hall2009weka}.  
For the Random Forest we use 50 trees (\textit{-I 50}). Pruning is enabled for J48. Finally we run the Instance Based classifier with \textit{-I -K 10} settings. 

\subsection{Features}
\label{sec:features}

Prior work on stance classification investigated various features  which can be categorized into linguistic, message-based, and topic-based categories \cite{mendoza2010twitter,qazvinian2011rumor,hamidian2015rumor,liu2015real,zeng2016unconfirmed,lukasik2016hawkes,zubiaga2016stance}. The following list summarizes the features adopted in this work.

\begin{compactitem} 
  \item \textbf{BOW (Bag of words):} For this feature we first create a dictionary from all the tweets in the out-of-domain dataset. Next each tweet is assigned the words in the dictionary as features. For words occurring in the tweet the feature values are set to the number of times they occur in the tweet. For all other words ``0'' is used.
  \item \textbf{Brown Cluster:} Brown clustering is a hard hierarchical clustering method and we use it to cluster words in hierarchies. It clusters words based on maximising the probability of the words under the bigram language model, where words are generated based on their clusters \cite{liang2005semi}. In previous work, it has been shown that Brown clusters yield better performance than directly using the BOW features \cite{lukasik2015classifying}. Brown clusters are obtained from a bigger tweet corpus that entails assignments of words to brown cluster ids. We used 1000 clusters, i.e. there are 1000 cluster ids. All 1000 ids are used as features however only, ids that cover words in the tweet are assigned a feature value ``1''. All other cluster id feature values are set to ``0''.   
\item \textbf{POS tag:} The BOW feature captures the actual words and is domain dependent. To create a feature that is not domain dependent, we added Part of Speech (POS) tags as additional feature. Similar to the BOW feature we created a dictionary of POS tags from the entire corpus (excluding the health data) and used this dictionary to label each tweet with it -- binary, i.e. whether a POS tag is present.\footnote{We also experimented with frequencies of POS tags, i.e. counting how many times a particular POS tag occurs in the tweet. The counts then have been normalized using mean and standard deviation. However, the frequency based POS feature negatively affected classification accuracy, so it has been omitted from the feature set.} However, instead of using just single POS tags, we created sequences containing bi-gram, tri-gram and 4-gram POS tags. Feature values are the frequencies of POS tag sequences occurring in the tweet. 
  \item \textbf{Sentiment:} This is another domain-independent feature. Sentiment analysis reveals the sentimental polarity of the tweet such as whether it is positive or negative. We used the Stanford sentiment\cite{socher2013recursive} tool to create this feature. The tool returns a range from 0 to 4 with 0 indicating ``very negative'' and 4 ``very positive''. First, we used this as a categorical feature but turning it to a numeric feature gave us better performance. Thus each tweet is assigned a sentiment feature whose value varies from 0 to 4.
  \item \textbf{NE:} Named entity (NE) is also domain independent. We check for each tweet whether it contains \textit{Person, Organization, Date, Location} and \textit{Money} tags and for each tag present, ``1'' is added, or a ``0'' otherwise. 
  \item \textbf{Reply:} This is a binary feature, which is assigned ``1'' if the tweet is a reply to a previous one, or a ``0'' otherwise. Tweet reply information is extracted from the tweet metadata. Again this feature is domain independent.
\item \textbf{Emoticon:} We created a dictionary of emoticons using Wikipedia\footnote{https://en.wikipedia.org/wiki/List\_of\_emoticons}. In Wikipedia those emoticons are grouped by categories, which we use as a feature. If any emoticon from a category occurs in the tweet, we assign for that category feature the value ``1'' -- otherwise ``0''. Again similar to the previous features this feature is domain independent.
\item \textbf{URL:} This is again domain independent. We assign the tweet ``1'' if it contains any URL, or ``0'' otherwise. 
  \item \textbf{Mood:} Mood detection analyses textual content using different view points or angles. Mood detection is performed using the tool from \cite{celli2016mood}, which analyses tweets from five different angles: amused, disappointed, indignant, satisfied and worried. For each of this angles it returns a value from -1 to +1. We use the different angles as the mood features and the returned values as the feature value.
%  \item \textbf{LIWC:} Use of LIWC dictionaries similar to related work discussed in Section \ref{sec:relatedWork}. 

\item \textbf{Originality score}: This is the count of tweets the user has produced, i.e. ``statuses count'' in the Twitter API.
\item \textbf{isUserVerified(0-1)}: Whether the user is verified or not.
\item \textbf{NumberOfFollowers}: Number of followers the user has.
\item \textbf{Role score}: This is the ratio between the number of followers and followees (i.e. NumberOfFollowers/NumberOfFollowees).
\item \textbf{Engagement score}: the number of tweets divided by the number of days the user has been active (number of days since the user account creation till today).
\item \textbf{Favourites score}: The ``favourites count'' divided by the number of days the user has been active.
\item \textbf{HasGeoEnabled(0-1)}: User has enabled geo-location or not.
\item \textbf{HasDescription(0-1)}: User has description or not.
\item \textbf{LenghtOfDescription in words}: The number of words in the user description.
\item \textbf{averageNegation}: We determine using the Stanford parser \cite{chen2014fast} the dependency parse tree of the tweet, count the number of negation relation (``neg'') that appears between two terms and divide this by the number of total relations.
\item \textbf{hasNegation(0-1)}: Tweet has negation relationship or not.
\item \textbf{hasSlangOrCurseWord(0-1)}: A dictionary of key words\footnote{www.noswearing.com/dictionary} is used to determine the presence of slang or curse words in the tweet.
\item \textbf{hasGoogleBadWord(0-1)}: Same as above but the dictionary of slang words is obtained from Google.\footnote{http://fffff.at/googles-official-list-of-bad-words}
\item \textbf{hasAcronyms(0-1)}: The tweet is checked for presence of acronyms using a acronym dictionary.\footnote{www.netlingo.com/category/acronyms.php}
\item \textbf{averageWordLength}: Average length of words (sum of word character counts divided by number of words in each tweet).
\item \textbf{hasQuestionMark(0-1)}: The tweet has ``?'' or not.
\item \textbf{hasExclamationMark(0-1)}: The tweet has ``!'' or not.
\item \textbf{hasDotDotDot(0-1)}: Whether the tweet has ``...'' or not.
\item \textbf{numberOfQuestionMark}: Count of ``?'' in the tweet.
\item \textbf{NumberOfExclamationMark}: Count of ``!'' in the tweet.
\item \textbf{numberOfDotDotDot}: Count of ``...'' in the tweet.
\item \textbf{Binary regular expressions applied on each tweet}: .*(rumor?|debunk?).*, .*is (that|this|it) true.*, etc. In total there are 10 features covering regular expressions.
\end{compactitem}

This work extends the features above, with new additional problem-specific features (\textbf{AF features}). AF features score the level of confidence in a tweet. We compute scores for surprise (\textit{surpriseScore (SS)}), doubt (\textit{doubtScore (DS)}), certainty (\textit{noDoubtScore (NDS)}) and support (\textit{supportScore (SPS)}) towards rumourous tweets. For each of these features a list of typical words is collected. We use this list to compute a cumulative vector using word2Vec \cite{mikolov2013distributed}. 
For each word in the list, we obtain its word2Vec representation, add them together and finally divide the resulting vector by the number of words to obtain the cumulative vector. Similarly a cumulative vector is computed for the words in the tweet excluding acronyms, named entities and URLs. We use cosine to compute the angle between those two cumulative vectors to determine each of the scores. Our word embeddings comprise the vectors published by~\newcite{baroni2014don}. The full list of tweet confidence AF features is as follows: 

\begin{compactitem}
\item \textbf{surpriseScore (SS)}: cosine between embedding of tweet content and the list of surprise words, e.g. ``surprise'', ``wonder'', etc. 
\item \textbf{doubtScore (DS)}: cosine between embedding of tweet content and the list of doubt words, e.g. ``doubt'', ``uncertain'', etc. 
\item \textbf{noDoubtScore (NDS)}: cosine between embedding of tweet content and the list of certainty words, e.g.``surely'', ``sure'', etc.
\item \textbf{supportScore (SPS)}: cosine between embedding tweet content and the list of support words, e.g. ``support'', ``confirm'', etc.  

Furthermore, the following two AF features are included: 

\item \textbf{initialTweetSim (ITS)} captures tweets that tend to support rumours. Every rumour is initiated by a tweet. We compute the cosine similarity based on word2Vec of the tweet being classified to the first tweet in the rumour thread. If the tweet is just a simple re-retweet of the initial tweet, this is taken as an evidence that the tweet is supportive of that tweet. 

\item \textbf{isQuestion (IQ)} indicates whether a tweet starts with an interrogative. The feature is binary and aims to capture questioning tweets.
\end{compactitem}

%Each tweet feature described above is extracted along with its class label. These are used by the classifiers either for learning purposes when they are run in training mode or for prediction if they run in testing mode. The training and testing steps are outlined in the next Section.

\section{Experimental Setup}
\label{sec:exSettings}

\begin{table}
\footnotesize
\centering
\begin{tabular}{p{2.5cm}p{2cm}p{1.5cm}}
\toprule 
\bf Classifier &\bf All features &\bf w.o. AF\\
\midrule
Decision tree  &74.16&72.25\\
Random Forest  &\textbf{79.02}&76.54\\
IBk            & 75.59&73.02\\
%SVM (RBF)      &74.16
\midrule
Baseline-Turing& 78.4&--\\
\bottomrule
\end{tabular}
\caption{\label{tab:classifiersPerformance} Accuracy scores of different stance classifiers for the RumourEval dataset. The baseline is the best performing system in the SemEval evaluation \textbf{Turing}.}
\end{table}

\begin{table*}
\footnotesize
\centering
\begin{tabular}{p{2.5cm}p{2.3cm}p{2.3cm}p{2.3cm}p{2.3cm}p{2cm}p{1.5cm}p{1.3cm}}
\toprule 
\bf classifier &\bf Ottawa shooting &\bf Ferguson riots &\bf Charlie Hebdo &\bf Sydney siege &\bf macro mean\\
\midrule
%SVM & 59.29&68.53 (ns) &65.78 & 64.39&64.49\\\hline
IBk &70.31*  & 72.35&\textbf{78.33} (ref) &75.44  &74.10\\
Decision tree & \textbf{76.28} (ref)& \textbf{75.20} (ref)& 78.21 & \textbf{80.01} (ref)&\textbf{77.42} \\
Random Forest & 69.39* &69.16 &74.57 &74.49 &71.90\\
Baseline - GP &62.28&64.31&70.66&65.04&65.57\\
Baseline - HP &67.77&68.44&72.93&68.59&69.43\\
\bottomrule
\end{tabular}
\caption{\label{tab:classifiersPerformancePheme} Accuracy scores for different stance classifiers on the PHEME dataset. *  indicates a significant difference to (``ref'') scores for each column of the table respectively as indicated by the paired t-test with $p<0.001$.}
\end{table*}

\begin{table*}[t]
\footnotesize
\centering
\begin{tabular}{p{3cm}p{2.3cm}p{2.3cm}p{2.3cm}p{2.3cm}p{1.5cm}p{1.3cm}p{0.8cm}}
\toprule
\bf classifier &\bf Ottawa shooting &\bf Ferguson riots &\bf Charlie Hebdo &\bf Sydney siege &\bf macro mean\\
\midrule
IBk / AF &69.26  & 69.54&77.09 &73.28 &72.29\\
J48 / AF&75.62& 74.85 & 77.05 & 79.21 &76.68 \\
Random Forest / AF& 67.87 &68.31 &75.40 &72.57 &71.03\\
%GP &62.28&64.31&70.66&65.04&65.57\\\hline
%HP &67.77&68.44&72.93&68.59&69.43\\\hline
\bottomrule
\end{tabular}
\caption{\label{tab:classifiersPerformancePhemeAF} Accuracy scores of different stance classifiers on the PHEME dataset with AF features removed.}
\end{table*}

\begin{table}
\footnotesize
\centering
\begin{tabular}{p{3.2cm}p{2.2cm}}
\toprule
\bf Features &\bf Accuracy\\
\midrule
All features&\textbf{79.02}\\
All without AF &76.54\\ 
All without ITS&78.55\\
All without SS&77.59\\
All without SPS&78.16\\
All without DS&78.36\\
All without NDS&77.59\\
All without IQ&78.64\\
\bottomrule
\end{tabular}
\caption{\label{tab:classifiersPerformanceEachAF} Contribution of each AF feature. Accuracy scores are for the Random Forest classifier on RumourEval data set with each feature removed in turn.}
\end{table}

\subsection{Baselines}

On the RumourEval dataset we run different classifiers (see Section~\ref{sec:classifiers}). We compare the performance of these classifiers against the best-performing system from the RumourEval challenge, namely \textbf{Turing}~\cite{kochkina2017turing}.

We also run all the classifiers from the RumourEval dataset on the PHEME dataset. The results are compared against the following baseline systems reported on the PHEME dataset:

\begin{compactitem}
\item \textbf{Gaussian   Processes (GP)} reported by~\newcite{lukasik2015classifying}.
\item \textbf{Hawkes Processes (HP)} reported by~\newcite{lukasik2016hawkes}. HPs make use of both temporal and textual information of tweets.
\end{compactitem}

\subsection{Training-Testing Settings}

We have two different settings. In the first setting we use the SemEval training data to train the models and apply on the testing data. In the second setting we perform training and testing on the PHEME dataset. For the PHEME dataset, we follow the leave one out (LOO) strategy taken by Lukasik et al. \shortcite{lukasik2016hawkes} to construct the training and testing data. In LOO n-1 rumours (all tweets within these rumours) are used for training and the resulting model is tested on the n$^{th}$ rumour. Finally, results are macro-averaged.

\section{Results}
\label{sec:results}

As shown in Table~\ref{tab:classifiersPerformance} (column two of the table) the best performing learner on the RumourEval dataset is the Random Forest classifier. It achieves the accuracy of \textit{79.02}, higher than any participating system in the RumourEval Task A.\footnote{The results are reported in \url{http://alt.qcri.org/semeval2017/task8/index.php?id=results}}

The results on the PHEME dataset are shown in Table \ref{tab:classifiersPerformancePheme}. Overall the best performing classifier is the J48 decision tree learner. The difference in accuracy scores between the classfiers is tested for significance using paired t-test (p$<$0.001). J48 is only significantly better than IBk and J48 for the \textit{Ottawa shooting} event type. In the remaining event types, J48 performs better, but not significantly better than IBk and Random Forest. 

All classifiers J48, IBk and Random Forest, however, outperform the \textbf{GP} and \textbf{HP} baselines on all event types \footnote{Although the significance test could not be run for the baselines, as the single data point values are not available, the proportion of difference in accuracy and the fact that it is the same data sets let us assume that the classifiers significantly outperform the baselines.}. 

What these results demonstrate is that simpler classifiers, such as J48 and Random Forest can outperform significantly more sophisticated machine learning methods (GPs and HPs in this case, and LSTMs in the RumourEval case),  thanks to the additional knowledge captured in the rich feature set. In contrast, for example, the GP and HP models relied primarily on BOW and Brown clustering features.  

%For each event type we used the best performing system and tested the significance of  using paired t-test to the other systems. The results show that the best performing system always significantly outperforms the GP and HP baselines (p$<$0.001).   

%-- can we also give a comparison on the different categories in the Pheme datasets? This is to be compared with Lukasik 2015 (gaussian processes)~\cite{lukasik2015classifying}
%KLB: Actually we should be comparing against Michal's ACL 2016 paper (hawkes processes IIRR), as this outperforms the GPs. There are similar tables in the ACL'2016 paper too. \cite{lukasik2016hawkes}
%DONE

\section{Feature Analysis}
\label{sec:analysis}
%\subsection{Feature Analysis}

The results described in Section~\ref{sec:results} are based on features reported by related work, enhanced by us with AF features (Section~\ref{sec:features}). We repeat the experiments with AF features removed from the feature set, in order to quantify the extent of their contribution. 

For the RumourEval dataset the results are shown in column 3 of Table~\ref{tab:classifiersPerformance}. The omission of the AF features leads to a performance decrease for all classifiers. The accuracy scores also fall below that of the SemEval winner \textbf{Turing} -- the state-of-the-art system on the RumourEval dataset. 

The results on the PHEME dataset are shown in Table~\ref{tab:classifiersPerformancePhemeAF}. The exclusion of the AF features leads to an overall drop in performance when compared to the same classifiers in Table \ref{tab:classifiersPerformancePheme}. However, these differences are not significant and the classifiers with AF features removed still perform at least as well as the GP and HP baselines (for the event type \textit{Ferguson riots}), or outperform the baselines (for all other event types). 

Table~\ref{tab:classifiersPerformanceEachAF} shows the accuracy scores of the Random Forest stance classifier, the best performing system on the RumourEval dataset when each AF feature is removed in turn. The results indicate that each AF feature contributes to the accuracy boost in stance classification. The highest accuracy loss results from removing the surprise (SS) and certainty (NDS) scores and the least -- when the \textit{isQuestion} (IQ) feature is removed. None of the AF feature removals cause a significant drop in accuracy. However, the loss is significant ($p<0.0001$) when all AF features are removed.

Both RumourEval and PHEME dataset evaluations show that the AF features play an important role in terms of achieving higher accuracy for tweet-based stance classification. They also show the importance of task or problem-specific feature engineering and point out that it is possible with some feature engineering effort to outperform state-of-the-art techniques that are typically considered more powerful and  sophisticated than traditional learning methods.  
%compare with tables 2 and 3 in~\cite{zubiaga2016stance}

%\subsection{Problem Complexity}

%One explanation for the success of our approach
%Measure the Rademacher complexity of our feature set?

% \section{Discussion -- NOT sure with this!}
% \label{discussion}

% The results in Table \ref{tab:classifiersPerformance} and Table \ref{tab:classifiersPerformancePheme} show different pictures with respect to the performance of the classifiers. On the RumourEval dataset we observed the Random Forest classifier as the best learner whereas in the PHEME dataset the best learner is the J48 decision tree classifier. The question here is which result to use to make the final decision about the best performing classifier. The results in Table \ref{tab:classifiersPerformance} are based on one single split of the data into training and testing sets whereas in the PHEME dataset we have \textit{n} different splits. Since the RumourEval dataset is derived from the PHEME dataset by taking the entire dataset and performing a single split one could replicate this process \textit{n} times to obtain \textit{n} splits. This would results in the LOO settings used in the PHEME dataset evaluation. Because of this we think that the results shown in Table \ref{tab:classifiersPerformancePheme} are to consider as the final decisions about the different classifiers.   

\section{Conclusion}
\label{sec:conclusion}

This paper tackled the problem of stance classification of tweets towards rumours. In our approach we use a simple classification approach, combining common features reported by related studies with our novel AF features, to boost overall accuracy. Our results show that this approach leads to significantly better results on both RumourEval and PHEME datasets compared to current state-of-the-art systems. Furthermore, our results show that the omission of the AF features proposed in this work leads to significantly lower performance. Adding AF to the feature set causes our approach to outperform the best performing system on the RumourEval dataset. These results show the importance of task- or problem-oriented feature engineering.

The proposed features are content based and work on text level. In our future work we plan to investigate features that are able to capture communication behaviours between users. We also plan to apply stance information as a feature in rumour veracity classification. 

\iftrue
\section*{Acknowledgments}
This work was partially supported by the European Union funded COMRADES (grant agreement No. 687847) and PHEME projects (grant agreement No. 611223), as well as an EPSRC career acceleration fellowship (EP/I004327/1).
\fi

\bibliography{stance}
\bibliographystyle{emnlp_natbib}

\end{document}